%% file: main.tex
\documentclass[letterpaper, 10 pt, conference]{ieeeconf}  

\IEEEoverridecommandlockouts                              

\usepackage{graphicx}
\usepackage{amsmath}
\usepackage{amssymb}

\usepackage{hyperref}
\usepackage{cleveref}

\title{
\LARGE \bf
Social Cohesion in Autonomous Driving
}

\author{Nicholas C. Landolfi and Anca D. Dragan
\thanks{Department of Electrical Engineering and Computer Sciences,
        University of California, Berkeley, Berkeley 94720
        {\tt\small nclandolfi,anca@berkeley.edu}}%
}

\newcommand\R{{\mathbb{R}}}
\newcommand{\prg}[1]{\noindent\textbf{#1. }}

\newcommand{\norm}[2]{\lVert#1\rVert_{#2}} 

\newcommand{\sref}[1]{Sec. \ref{#1}}
\newcommand{\figref}[1]{Fig. \ref{#1}}
\DeclareMathOperator*{\argmax}{arg\,max}

\usepackage{color, colortbl}
\definecolor{Gray}{gray}{0.9}
\usepackage{array,multirow}
\usepackage{tabularx}

\renewcommand{\baselinestretch}{0.95}

\begin{document}

\maketitle

\input{abstract}
\input{intro.tex}
\input{driving.tex}
\input{algorithm.tex}
\input{analysis.tex}
\input{study.tex}
\input{discussion.tex}
\input{acknowledgements.tex}

\bibliographystyle{IEEEtran}
\bibliography{biblio} 

\end{document}

%% file: abstract.tex
\begin{abstract}
    Autonomous cars can perform poorly for many reasons. They may have perception issues, incorrect dynamics models, be unaware of obscure rules of human traffic systems, or follow certain rules too conservatively. Regardless of the exact failure mode of the car, often human drivers around the car are behaving correctly. For example, even if the car does not know that it should pull over when an ambulance races by, other humans on the road \emph{will} know and \emph{will} pull over. We propose to make \emph{socially cohesive} cars that leverage the behavior of nearby human drivers to act in ways that are safer and more socially acceptable. The simple intuition behind our algorithm is that if all the humans are consistently behaving in a particular way, then the autonomous car probably should too. We analyze the performance of our algorithm in a variety of scenarios and conduct a user study to assess people's attitudes towards socially cohesive cars. We find that people are surprisingly tolerant of mistakes that cohesive cars might make in order to get the benefits of driving in a car with a safer, or even just more socially acceptable behavior.
\end{abstract}

%% file: intro.tex
\section{Introduction}

Humans are remarkable drivers. Yes, we get fatigued, we get distracted, we drive when we should not. But we also are amazing at understanding what is going on around us. We understand what traffic cones mean, we know how to handle construction zones, we know to pull over when an ambulance passes. When we switch cities or even countries, we might be confused for a while, but we easily adapt to the new driving culture and to the unwritten rules of the road. 

Autonomous cars will \emph{eventually} get there.  But in the meantime, we have to expect that they will be faced with situations they cannot yet handle correctly on their own. They might face a new kind of traffic cone. They might enter a construction area where they are supposed to drive on the lane markings and not in the center of the lane. They might be taken to a different city or even country, where people always drive at least 20 miles over the limit on the highway (or always 5 miles under, for that matter). They might have to avoid some obstacle they do not see or have not detected. 

All the above are very difficult cases to deal with. But typically, the car will not be driving in isolation: in the near-term, there will be other, human drivers around. They will be navigating these complex situations with ease.

\begin{quote}
\emph{
	Our insight is that there is implicit information in the behavior of human-driven vehicles that autonomous cars can leverage to be safer and more socially acceptable.
}
\end{quote}

One option for leveraging this information might be to have a rich enough hypothesis space for how the world may differ from the car's understanding (e.g., where all hidden obstacles might be), and treat human driver trajectories as evidence about this hidden variable. That way, when the robot sees a person swerve to the next lane, it would deduce that the person is probably avoiding some obstacle, even though the robot has not detected this obstacle.

Unfortunately, constructing a hypothesis space rich enough to cover all possibilities might be difficult. When the car sees an ambulance coming, and sees everyone moving to the right lane, it may not have anything in its hypothesis space to explain that behavior -- it either knows about the ambulance rule or it does not. 

In this work, we take a simpler approach to leveraging human driver information. We propose cars that are \emph{socially cohesive}: if everyone else is doing it, then our car should do it too. If everyone else is driving 10 miles over or 5 miles under the speed limit, so should the autonomous car. If everyone else is avoiding something in a lane, so should the autonomous car, even if it does not detect any obstacle. If everyone is driving on the lane divider as opposed to in the center of the lane because it is a construction zone, then so should the autonomous car. 

Although there might be aspects of the real world that will still be hard to understand for cars in the near future, we think that being able to follow along can help them overcome some of their limitations. For instance, imagine the recent accident where an autopilot-driven car ran into a truck because its perception system did not detect it \cite{teslacrash}. If the car had followed everyone else on the road who was avoiding the truck, it would have been safe. Our contributions are three-fold:

\begin{figure}
\includegraphics[width=\columnwidth]{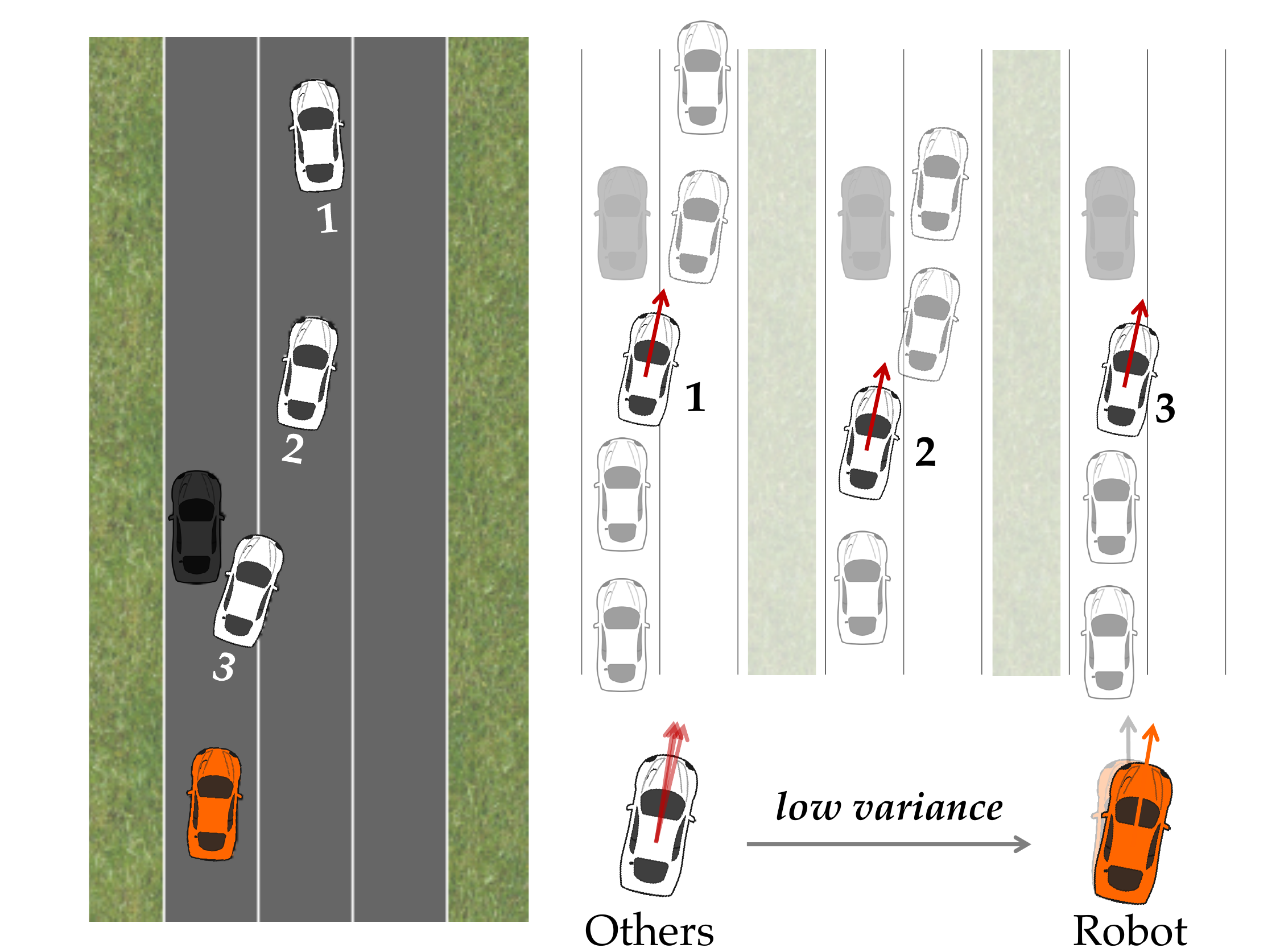}
\caption{Three cars (white) in front of the robot car (orange) swerve. The robot car can not see the stalled (gray) car, but can see the low variance in the white cars' trajectories. We incentivize the robot to be cohesive, to also swerve, when it sees agreement in the behavior of other drivers.}
\label{fig:frontfig}
\vspace{-0.5cm}
\end{figure}


\prg{1. Algorithm for socially cohesive autonomous driving} 

Following everyone else sounds deceptively simple. In reality, the car must decide \emph{what aspects} to follow (e.g., to only match the speed or follow the full trajectory of other cars), \emph{who} to follow (e.g., only cars in its lane, or all cars around it), and \emph{when} to follow. Our idea is to use \emph{variance} in other car's behavior as a natural way to answer these questions. Our car only follows aspects with low variance (i.e., that are consistent across other cars), for only the subset of cars that exhibits the low variance, and in proportion to how low the variance is: when drivers are all doing different things, there is no social cohesion, and our car will not be socially cohesive; but when drivers are consistently swerving, moving over, or going above the speed limit, our car will follow along. 

\prg{2. Analysis of our algorithm across diverse scenarios}

We test this algorithm in different scenarios and show that the same approach can reliably pick up on the right aspects for cohesion and not imitate the wrong ones. Of course, false positives still occur -- if enough cars reliably turn right or take an exit, social cohesion is going to incentivize the car to do the same. In some cases, that is good, because likely there is something (like a construction setup that the robot did not detect) causing everyone to make the same decision. But in other cases, it is just coincidence, and the car gets tricked by that into following along. Social cohesion is not a silver bullet, and there will definitely be situations where it leads to undesired consequences. 

\prg{3. Exploration of human reactions to cohesive cars}

Social cohesion has the benefit of increasing safety and social acceptance in many driving scenarios, with the trade-off that it will also mistakenly follow unintended behavior in others. We run two users studies to evaluate people's attitudes towards a cohesive car. We find that, unsurprisingly, people prefer the cohesive car when they see that it can help prevent collisions, and are willing to tolerate mistakes like taking the wrong exit. Surprisingly, even when we remove the collision scenario, and social cohesion is left with primarily social advantages, we find that participants still show a significant preference for the cohesive car, both in the hypothetical role of other drivers as well as in the role of passengers. 

Of course, social cohesion is no permanent solution -- the goal of autonomous driving is to reduce human-driven vehicles on the road, so there may not be humans to be socially cohesive with (though it might be good to be socially cohesive to other autonomous cars if there is no direct communication). But what social cohesion can do is perhaps get cars on the road faster.

%% file: driving.tex
\section{Formulation of Driving}

Following prior work \cite{sadigh-human-rss2016,sadigh-iros2016}, we specify the behavior of our autonomous car via a reward function, which depends on the state and control sequences of our robot. 
We construct the robot's reward at each time step from a linear combination of features. 
The autonomous car will maximize this reward over a finite horizon.
We assume known, deterministic and discrete dynamics. 

\subsection{Notation}
Let $n$ be the dimension of the state space and $m$ the dimension of the control space. 
Let $x^t \in \R^n$ denote the state vector at a particular time $t$ and likewise $u^t \in \R^m$ denote the control.
Let $x^0 \in \R^n$ denote the robot's current state.
Let $f: \R^n \times \R^m \to \R^n$ denote the transition function.
Each subsequent state is found via the transition function: $x^{t+1} = f(x^{t}, u^{t})$.

Let $p$ be the number of features. 
Let $\phi: \R^n \times \R^m \to \R^p$ denote the function which computes the features for a particular state and control pair. 
Let $\theta \in \R^p$ be the reward parameters (i.e., the weights in the linear combination of features).
Let $T$ denote the robot's horizon.
Let $\mathbf{u} = (u^0, \ldots, u^T)$ be the finite sequence of robot controls.

\subsection{Model Predictive Control (MPC)}
\label{sec:mpc}

\prg{Finite Time Horizon with Re-planning}
We use model predictive control \cite{morari1993model} to generate the actions for our car.
At each time step we optimize:
\begin{equation}
    \mathbf{u}^* = \argmax_{\mathbf{u}} \left\{ \sum_{t = 0}^{T} \theta^\top\phi(x^t, u^t) \right\}.
    \label{eq:ustarfin}
\end{equation}
Our car applies the first control of the sequence $\mathbf{u}^*$ and then re-plans again at the next time step.

\prg{Solution via Quasi-Newton Optimization} 
Given $\theta$, $\phi$ and $x^0$ we locally solve \eqref{eq:ustarfin} using gradient-based methods. 
Our formulation does not depend on a particular optimization method. 
For our analysis in \sref{cases} we use L-BFGS \cite{andrew2007scalable}.

\subsection{Feature Map \& Reward Parameters}

\begin{figure}
\includegraphics[scale=0.15]{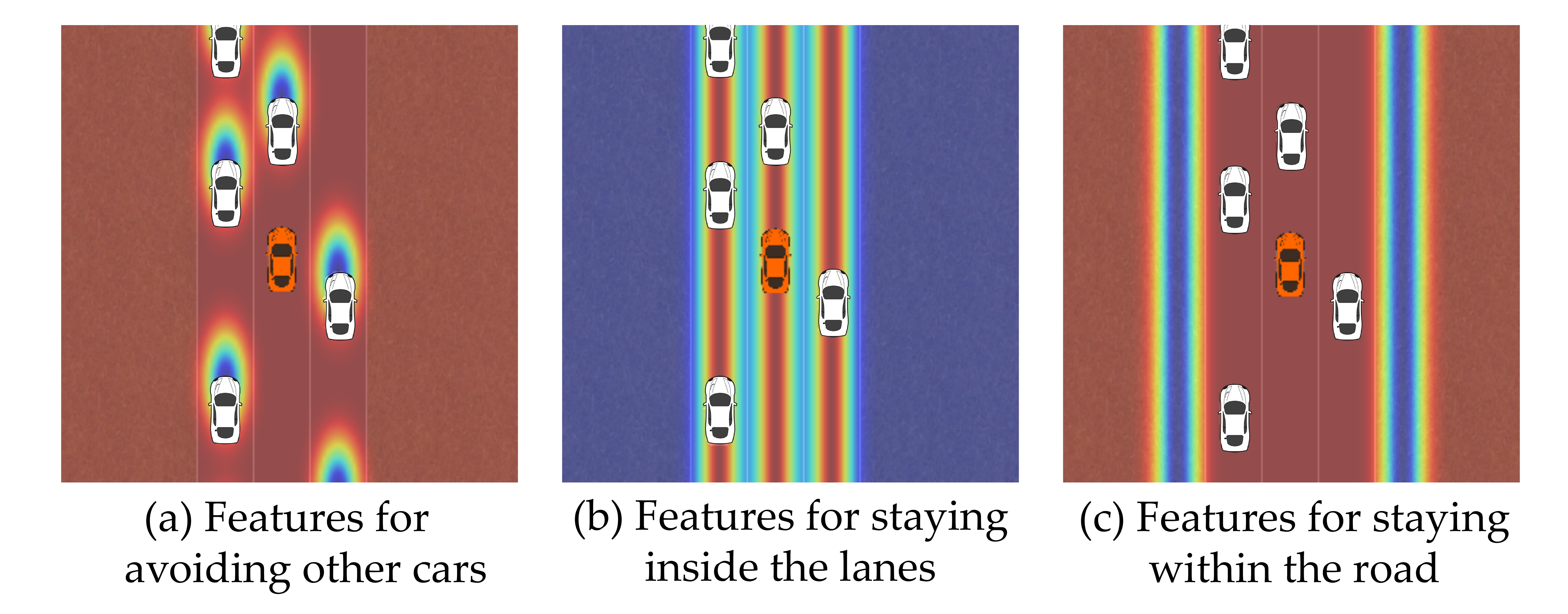}
\caption{\textbf{Driving Features.} Red corresponds to higher reward, blue to lower reward. The map $\phi$ contains features for avoiding collision with other cars (a), holding a particular lane (b) and staying within the road boundaries (c). These and other features are weighted by $\theta$ in the reward.}
\label{fig:features}
\vspace{-.5cm}
\end{figure}

Prior work \cite{sadigh-human-rss2016} inspires our features. We visualize examples of these in \figref{fig:features}. In addition to those features shown for the road, lanes and other cars, we include features for the magnitude of the control inputs. Collectively, these features enable our car to effectively drive on the road.

To find a reasonable value for $\theta$, we collect examples of driving in a simulator. 
We then find $\theta$ via inverse reinforcement learning \cite{ng2000algorithms,ziebart2008maximum}, which is discussed for the driving case in \cite{sadigh-human-rss2016,shimosaka2014modeling,kuderer2015learning}. As in \cite{sadigh-human-rss2016} we handle the continuous state and action spaces using continuous inverse optimal control with locally optimal examples \cite{levine2012continuous}. 

%% file: algorithm.tex
\section{Algorithm for Social Cohesion}

To be socially cohesive, the car needs to answer \emph{who} and what to imitate as well as \emph{when} and \emph{how} to do it.

\subsection{Overall Approach}

We will answer the \emph{what} by devising features on the trajectories of other cars; this is the structure we consider imitating. 
We will answer the \emph{who} by making groups with which the car might want to be cohesive (e.g., everyone on the road or everyone in the lane). 
We will answer the \emph{when} based on the sample \emph{variance} of these features within each group: the car imitates a feature of a group when that feature has low enough variance. 
This brings us to the \emph{how}. 
\subsection{Social Features}

\begin{figure}
    \includegraphics[scale=0.35]{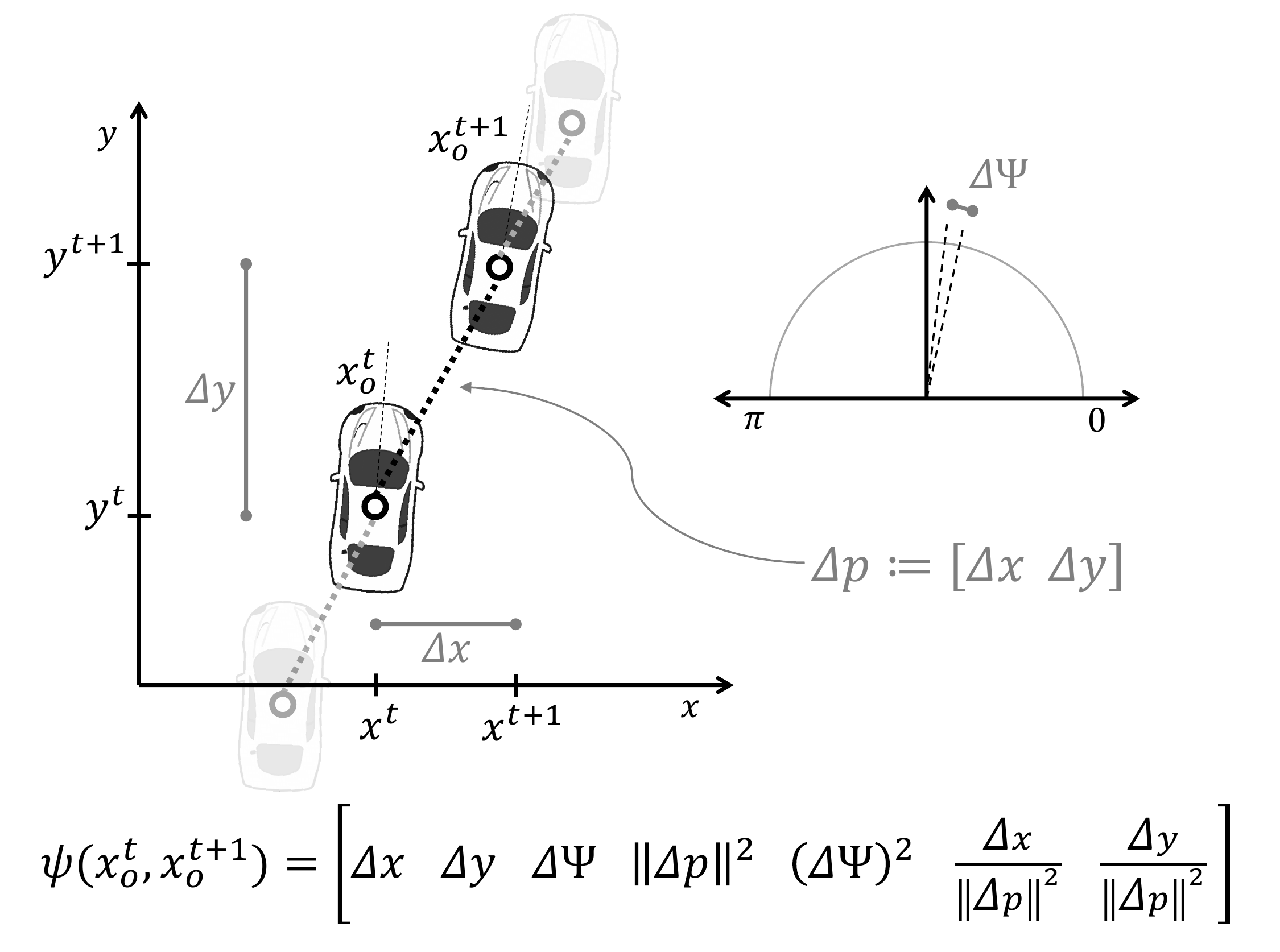}
    
    \caption{\textbf{Social Features.}  Defined pairwise in time on the states of the other cars. They capture change in position and orientation. We normalize these differences to capture direction, handling the case of multiple drivers moving in the same direction at different speeds. Let $x_o^t$ and $x_o^{t+1}$ denote two sequential states of another car's trajectory.}
    
    \label{fig:socialfeatures}
    
    \vspace{-0.5cm}
\end{figure}

We compute social features pairwise in time on the states of the trajectories of other cars  (visualized in \figref{fig:socialfeatures}). 
Let $k$ be the number of social features. 
Let $\psi: \R^n \times \R^n \to \R^k$ denote the function which computes these features.

\subsection{Cohesion as Matching Social Features}

For each feature $\psi_i$ and group $j$ we compute the sample mean $\mu_{i,j}$ and the sample variance $\sigma_{i,j}^2$. 
We want the car to match the mean of feature $i$ in group $j$ when this feature has \emph{low} variance (i.e., is cohesive), and ignore it when it has \emph{high} variance. 

Further, it does not suffice to match the mean: that will make the car socially cohesive, but it might not make it usable -- the feature might be merely one aspect of behavior, not a full specification. The car might then match the speed of other cars, for instance, but it sill needs to reach its own destination and avoid collisions! 

Therefore, our idea is for the car to \emph{trade off} between its original reward and a social cohesion reward, where \emph{the trade-off value depends on the variance of the cohesive feature}. This seamlessly achieves our goals: it ensures that the car is only cohesive when there is a cohesive feature on the road, it still optimizes its objective in the absence of matching that feature, and it smoothly interpolates to not caring about what others are doing when others are not cohesive. 

\subsection{Cohesion-Augmented Reward}
Let $M$ be the number of groups.
Let $d$ be a metric.
Let $\beta$ denote the parameter controlling the trade-off between the original reward and the cohesion reward.
At every step, we optimize a cohesion-augmented reward:

\begin{equation}
    R_c := 
    \underbrace{\sum_{t = 0}^{T} \theta^\top\phi(x^t, u^t)}_{\text{original}} - \beta\underbrace{
    \sum_{i=1}^{k}\sum_{j=1}^{M} \frac{d(\mu_{i,j}, \psi_i(x^0, x^1))}{\sigma_{i, j}^2}
    }_{\text{cohesion}}.
    \label{eq:cohesr}
\end{equation}

The variance of feature $i$ in group $j$ scales the distance (according to the metric $d$) of the social features computed on the first two states of our car's planned trajectory ($x^0$ and $x^1$) and $\mu_{i, j}$. 
Notice that adding a social cohesion term to the objective can be done across a variety of planners. 
We perform MPC as in \sref{sec:mpc} with the new reward $R_c$.

\subsection{Grouping Features}
\label{sec:groupingfeatures}

If we looked at social features across all cars on the road, we might miss cohesive behavior only occurring on some parts of the road, or behavior only displayed by cars at a specific position. 
Thus our feature groups depend on the \emph{state} of the car generating the trajectory and the \emph{time} at which the feature was generated.

For example, our analysis in \sref{cases} considers three feature groups: (1) those generated by all cars which were in the robot's current \textit{position} at any time, (2) those generated by all cars which are in the robot's \textit{lane} at the present time, and (3) those generated by all cars which were near the robot on the \textit{road} at the present time.

%% file: analysis.tex
\section{Analysis}
\label{cases}

\begin{figure*}
	\includegraphics[width=\textwidth]{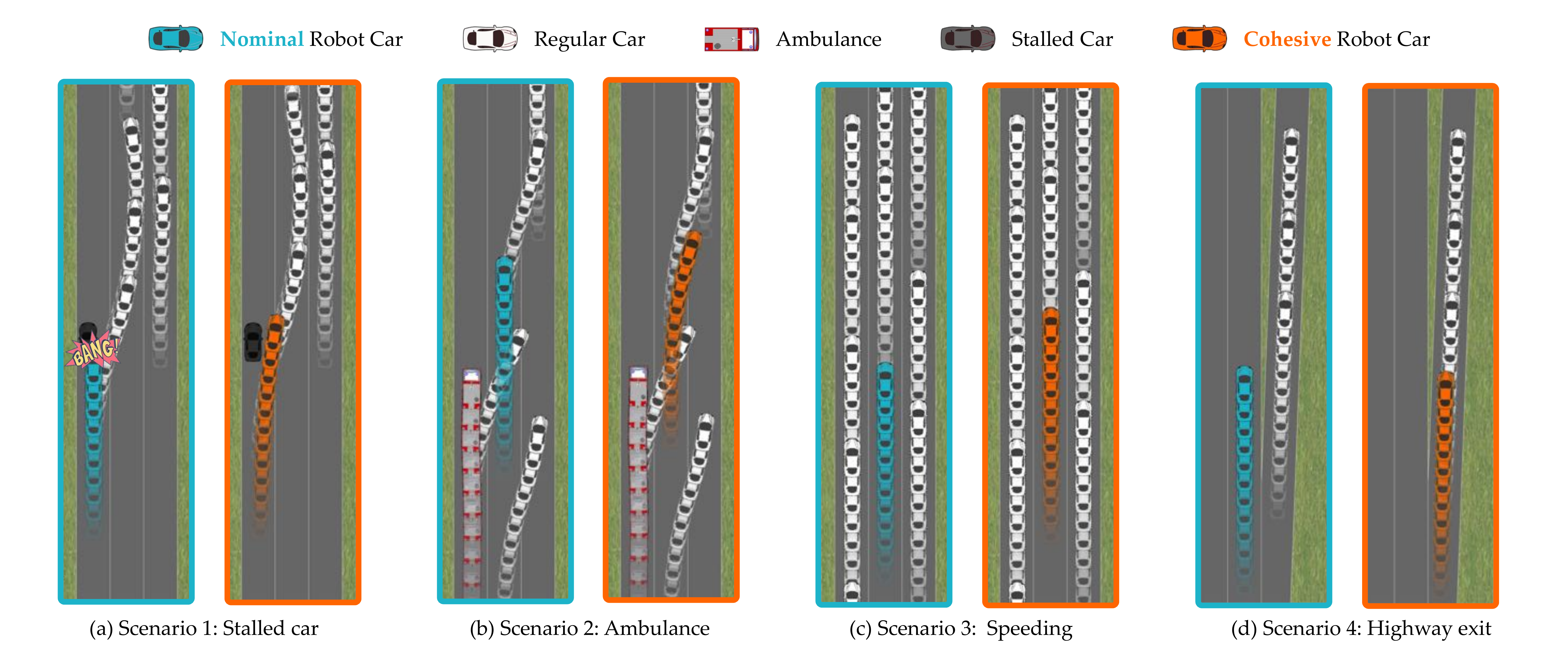}	
	\caption{\textbf{Driving Scenarios.} In (a), all white cars swerve around a hard-to-see stalled car, and the cohesive car swerves as well. In (b), an ambulance emerges in the far left lane, all white cars know to move to the right lane, and the cohesive car notices this behavior. In (c), all white cars are traveling above the speed limit, and the cohesive car learns it can go faster. In (d), all white cars take the exit, and the cohesive car is tricked into taking the exit. None of these plans use hand-coded strategies. The behavior \emph{emerges} out of optimizing the cohesive reward function. By capturing cohesion in the reward, our algorithm is able to generalize to these situations, enabling the robot to adapt.}
	\label{fig:scenarios_hri}
	\vspace{-0.5cm}
\end{figure*}

In this section we introduce four scenarios highlighting the benefits and trade-offs of cohesive planning.
In \sref{sec:study1} we perform a study to understand how people react to these discrepancies in behavior.
We visualize the scenarios in \figref{fig:scenarios_hri}. 
We depict human cars in white. 
We depict the robot cars in blue and orange, depending on the reward used.
The blue robot plans using the reward in \eqref{eq:ustarfin}.
The orange robot plans using the reward in \eqref{eq:cohesr}.
The gray car in Scenario 4 is stalled. 

\prg{Conditions for Analysis Across Scenarios} 
Each scenario considered has a fixed initial configuration for the cars.
In the control condition (blue), we have a \emph{nominal} car plan with the regular reward; \emph{avoid obstacles and stay on the road}. 
In the experimental condition (orange), we have a \emph{cohesive} car plan with the cohesive reward; drive well and \emph{match social features that have low variance}. 
In the sections following our discussion of implementation, we analyze how and why augmenting the reward with a social cohesion term impacts the behavior.

\subsection{Implementation}
\label{sec:sim}

\prg{Driving Simulation}
We build on the simulator used in \cite{sadigh-human-rss2016}. 
The car dynamics are point mass. 
Let $x$ and $y$ denote the coordinates of a car in the plane. 
Let $\Psi$ denote the car's orientation.
Let $s$ denote the car's speed. 
Then the car's four-dimensional state is $(x \enskip y \enskip \Psi \enskip s)^\top$.
Let $\Theta$ denote the steering and $a$ denote the acceleration of the car.
Then the car's two-dimensional control is $(\Theta \enskip a)^\top$.
Let $\mu$ denote the coefficient of linear friction experienced by the point masses.
 We use the discrete form the continuous dynamics,
\begin{equation}
    (\dot{x} \enskip \dot{y} \enskip \dot{\Psi} \enskip \dot{s}) = (s\cdot\cos(\Psi) \enskip s\cdot\sin(\Psi) \enskip s \cdot \Theta \quad a-\mu\cdot s).
    \label{eq:actualdynamics}
\end{equation}

\prg{Social Cohesion Algorithm} 
We define the social cohesion map $\psi$ on a pair of physical states as in \figref{fig:socialfeatures}.
We use the three groups discussed in \sref{sec:groupingfeatures} to split these social features. Group 1 contains social features corresponding to states at the robot's \textit{current position, at all times}. Group 2 contains social features corresponding to states which are in the robot's \textit{current lane, at the present time}. Group 3 contains social features for all cars \textit{nearby on the road, at the present time}. 
For simplicity, we set $\beta=1$; the cohesive reward will dominate the original reward in the cases of strong cohesion\footnote{An implementation detail: one must take care to correctly scale the feature values, so that the variances between features are comparable. For the case of differences discussed here, we normalize each feature by its maximum difference expected. This process avoids comparing a variance in radians with a variance in meters.}.
We use the software package Theano \cite{bergstraal:2010-scipy,Bastien-Theano-2012} to symbolically compute all Jacobians and Hessians and use the L-BFGS \cite{Byrd94alimited-memory} implementation in SciPy \cite{scipyguys}.

\subsection{Social Cohesion Across Diverse Scenarios}
\label{sec:scenarios}

Deciding \emph{what} to follow varies across scenarios. In some instances, as in the stalled example, it depends on what cars in a particular lane and position are doing. Other times, as in the ambulance example, the cohesive car should emulate the full trajectories of all the cars on the road. In still other cases, as in the speeding example, we want to mimic all cars on the road but only one aspect of their behavior.

\prg{Scenario 1: Stalled car}
The autonomous car approaches a \emph{stalled} (gray) car.
We model the robot car as \textit{unable to see the obstacle}, which could occur if the vision system failed.
However, this example also models a lane closure or the presence of ice on that spot of the road. The car may not be equipped to fully notice and avoid these latter two effects. 
The baseline car, failing to see the obstacle, crashes. 
In contrast, the cohesive car notices that \emph{all of the other cars swerve} at that location, and so \emph{also swerves}.

\prg{Scenario 2: Ambulance}
An ambulance approaches in the far left lane. On one hand, the autonomous car does not know that it should pull over to the right lane to allow the ambulance clear passage. On the other hand, all of the human cars know that they should do so, and begin executing trajectories heading toward the right lane.
The cohesive car notices the high agreement in trajectories among the cars on the entire road, and begins to emulate the behavior. The baseline car does consider the other cars beyond treating them as obstacles when planning and therefore fails to yield to the emergency vehicle. 

\prg{Scenario 3: Speeding}
The autonomous car starts by traveling the speed limit while all other cars on the road are traveling slightly \emph{above the speed limit}. 
The baseline car does not look for this cohesion in speed, but the cohesive car sees the high agreement and therefore \textit{begins to drive faster because everyone else is} doing so. 
Of course, it never fully reaches the speed of the other cars because its original objective was to travel the speed limit. The baseline car maintains the speed limit, potentially causing traffic. 

\prg{Scenario 4: Highway exit}
The autonomous car drives behind three human cars in a lane which becomes a highway exit.
We assume that the fastest route to the autonomous car's destination is to remain on the highway.
However, when all three cars in its lane exit, the robot car \textit{mistakenly also takes the exit}.
There is strong agreement in the trajectories of all the cars in the lane, and therefore other cars on the road are extraneous to this effect.
If this scenario actually occurred, it may be slightly ambiguous as to whether there exists some unseen obstacle ahead, but we analyzed this case because it is fairly common for several cars ahead of you in your lane to follow similar trajectories which are different from your optimal one. 
This issue might be alleviated by a more sophisticated selection of $\beta$, which is the subject of future work as mentioned later in \sref{sec:limitations}.

\subsection{Sensitivity Analysis}
We have seen the cohesive car handle a variety of scenarios, but we should check that its ability to pick up on this structure is not a coincidence. 
The car needs to be able to determine \emph{when} to follow, because it should not \textit{always} follow other cars on the road. 

\begin{figure}
\includegraphics[scale=0.35]{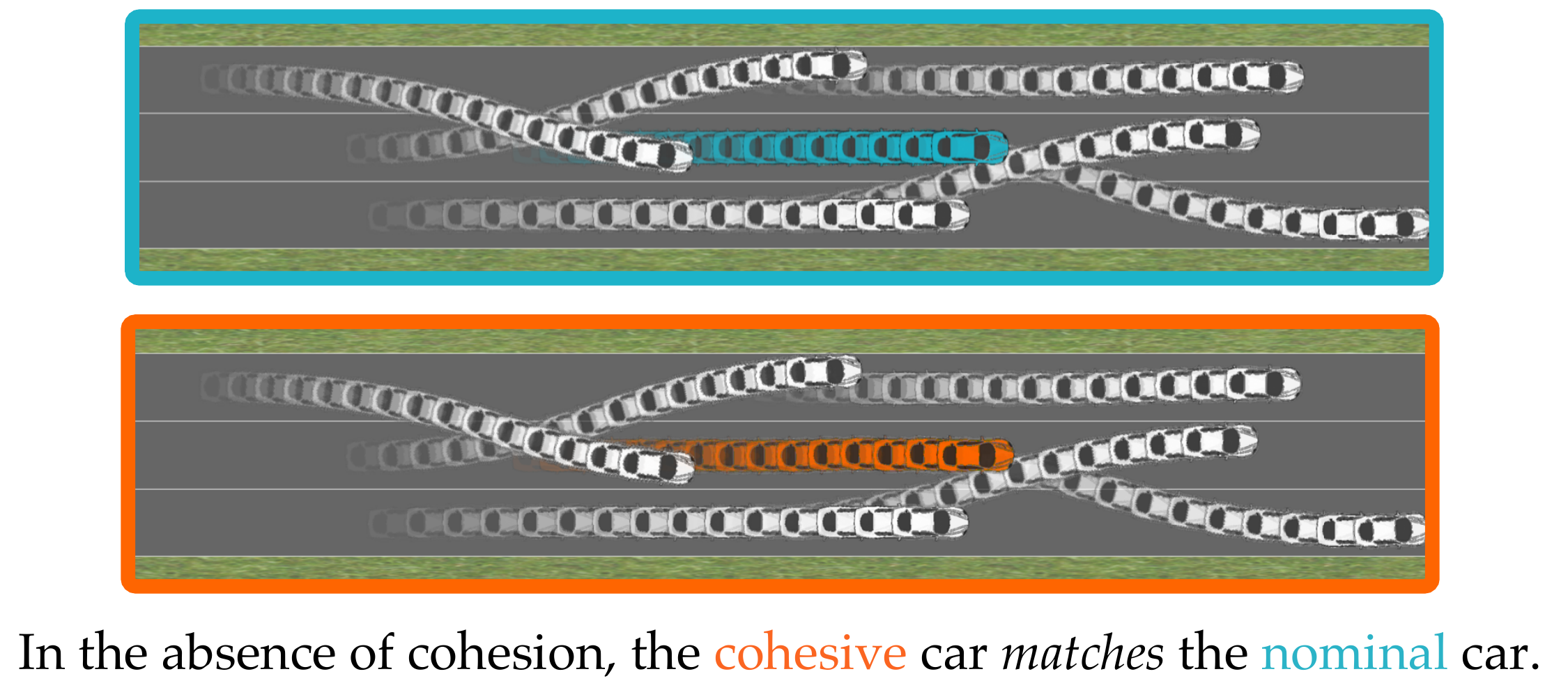}
\caption{\textbf{Nothing to match.} In the case that all the behavior of the cars is different,  both the cohesive and nominal car plan similarly.}
\label{fig:nothingtomatch}
\vspace{-.5cm}
\end{figure}

\prg{Nothing to Follow}
If there is no similarity in the behavior of other cars, then we should not follow them.
In \figref{fig:nothingtomatch}, we show that the cohesive car follows a near-identical trajectory to that of the nominal car, when everyone else on the road is behaving erratically in their trajectory and speed.

\begin{figure*}
\includegraphics[width=\textwidth]{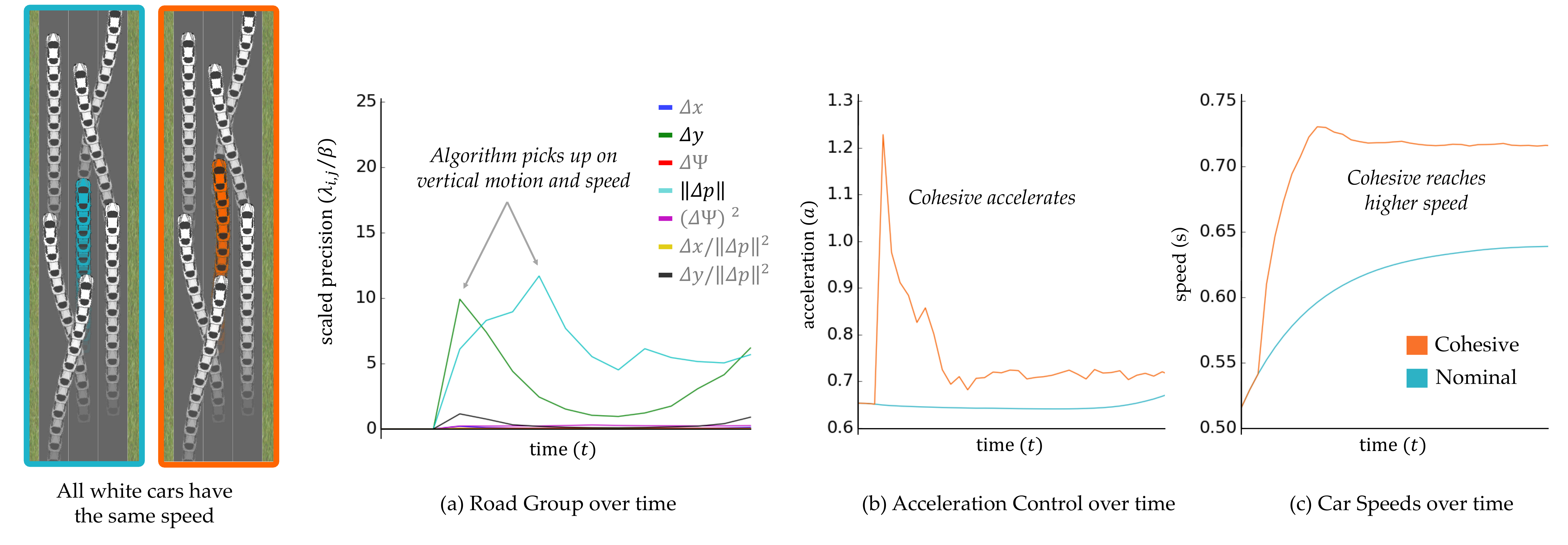}
\caption{\textbf{Identifying speed as a common feature}. We show the nominal and cohesive planners in a situation similar to Scenario 3, except now with drivers switching lanes around the robot car, as real drivers would on the highway. All other cars maintain the same speed which is larger than that of the robot car. Let $\lambda_{i,j} = 1/\sigma_{i,j}^2$ denote the sample precision. Plot (a) shows that the car identifies that the speed is similar across all cars on the road, and so the autonomous car accelerates. }
\label{fig:speeding}
\end{figure*}

\prg{Selecting Only the Correct Features} 
Even if there \textit{is} similarity in the behavior of other cars, our algorithm should only match the \textit{relevant features}.
For example, if everyone is going the same speed, but swerving different directions, we should only follow the speed, and not the direction.
Consider \figref{fig:speeding}, which depicts a situation similar to Scenario 3 in \figref{fig:scenarios_hri}. 
All cars are traveling faster than the robot car but each of the cars is following a distinct trajectory; the cars have little cohesion over direction.
This setup is identical to \figref{fig:nothingtomatch}, except all white cars travel at the \emph{same speed}.

Here, the plot of the sample precisions highlight the low variance among the features corresponding to magnitude of distance traveled $(\norm{\Delta p}{})$ and the distance traveled in the $y$ direction $(\Delta y)$; these are indeed the features which capture speed.
The cohesive car accelerates before reaching a steady speed closer to matching the other cars on the road.

\begin{figure}
\includegraphics[scale=0.35]{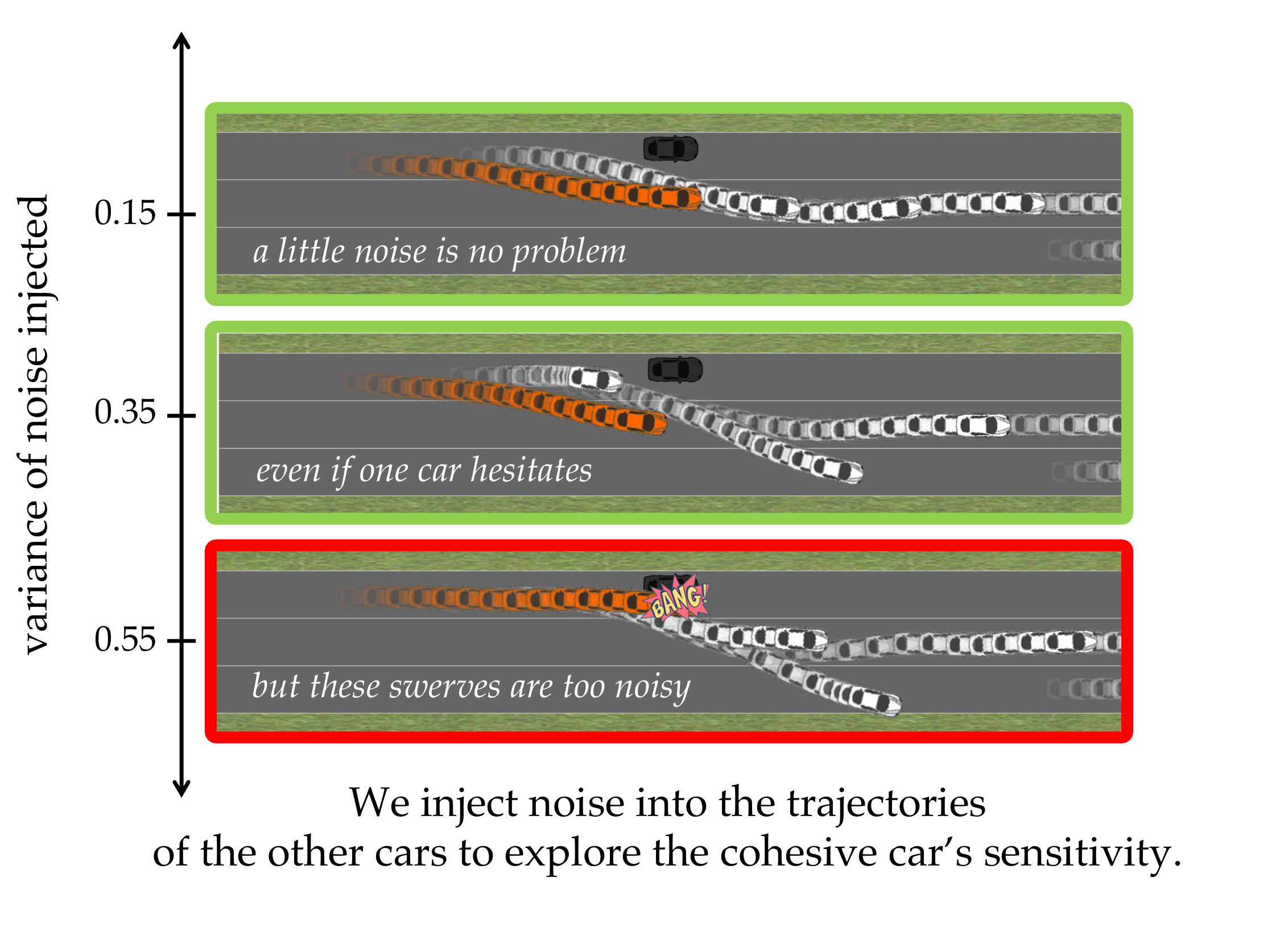}
\caption{\textbf{Noise in the Swerve.} In the top two panels, the cohesive car successfully completes the swerve, despite some noise in the trajectories of other cars. In the bottom panel, the other cars' behavior varies so much that the robot car is unable to discern that it should swerve, and crashes.}
\label{fig:sensitivity}
\vspace{-.5cm}
\end{figure}

\prg{Adding Noise to the Trajectories of Other Cars} Finally, we explore adding noise to the trajectories of the other cars to investigate the effects on the cohesive car's ability to distinguish features. We visualize these experiments in \figref{fig:sensitivity}. We find, of course, that the sample variances grow with the size of the noise injected. Until this noise causes non-trivial deviations in the $x$ coordinate of the others car, however, the cohesive car is still able to discern the swerve.

%% file: study.tex
\section{User Study}
\label{sec:study1}

\subsection{Experiment Design}
\label{sec:study1design}
We designed a survey largely focused on the perception of the participants when in the role of the passenger in the car, while also including a few items on the role as another driver sharing the road with the car. Table 1 shows our items, with scales focusing on the safety aspect (would passengers feel safe in the car), the social aspect (would passengers feel at ease with how other drivers are perceiving them), the frustration aspect (would passengers feel annoyed by the car's behavior), and overall passenger experience. We also included an item about willingness to share the road with the car, and whether the participant would judge the car's passengers. We showed participants videos of the four scenarios seen in \figref{fig:scenarios_hri}, with the same context as provided in \sref{sec:scenarios}.

\prg{Manipulated Factors} 
We changed the algorithm run by the autonomous car. In one case, the car uses the \emph{baseline planner}. In the other case, it uses the \emph{cohesive planner}.

\prg{Other Variables}
We showed four different scenarios, the same as those in \figref{fig:scenarios_hri}. For each scenario, we revealed the behavior of the nominal and cohesive cars. We counterbalanced the order of the algorithms.

\prg{Dependent Measures}
We designed eleven Likert scale questions to assess reactions to the cars. These ranged from the safety aspects of the car's behavior to social judgment felt by or toward a passenger of the car. These questions are listed in Table 1. We asked these questions after showing participants a scenario, prefixing the question with ``Based on its behavior in this one scenario, and forgetting about any previous scenarios..." After seeing all four scenarios, we asked the same questions, but this time prefixed by, ``Based on all the scenarios I've seen..." Lastly, we asked participants for a hard-choice preference at the end: which car would they prefer to ride in, and with which car would they prefer to share the road.


{\renewcommand{\arraystretch}{2.6}
\begin{table}
  \label{tab:likert}
  \begin{tabular}{clc}
    & \multicolumn{1}{c}{\textbf{Likert Questions}} & \multicolumn{1}{c}{\textbf{Cronbach's $\alpha$}}  \\
    \hline
    
    \parbox[t]{2mm}{\multirow{2}{*}{\rotatebox[origin=c]{90}{\textbf{safety}}}} 
    & \multicolumn{1}{p{5cm}}{Q1: I would be worried to ride in this car.} & \multirow{2}{*}{.83} \\
    & \multicolumn{1}{p{5cm}}{Q2: I would feel safe riding in this car.  } & \\
    \hline
    
    \parbox[t]{2mm}{\multirow{4.4}{*}{\rotatebox[origin=c]{90}{\textbf{passenger social}}}}
    & \multicolumn{1}{p{5cm}}{Q3: I would be embarrassed to ride in this car.} & \multirow{4.4}{*}{.82} \\
    & \multicolumn{1}{p{5cm}}{Q4: If I rode in this car, I would feel like others on the road would judge me.}  &\\
    & \multicolumn{1}{p{5cm}}{Q5: If I rode in this car, I would feel like others on the road would think highly of me.}  &\\
    \hline
    
    \parbox[t]{2mm}{\multirow{3.2}{*}{\rotatebox[origin=c]{90}{\textbf{frustration}}}} 
    & \multicolumn{1}{p{5cm}}{Q6: I would feel frustrated to ride in this car.}  & \multirow{3.2}{*}{.82} \\
    & \multicolumn{1}{p{5cm}}{Q7: If I were a driver in one of the white cars on the road, I would be annoyed by this car.}  & \\
    \hline
    
    \parbox[t]{2mm}{\multirow{2.5}{*}{\rotatebox[origin=c]{90}{\textbf{pass. exp.}}}} 
    & \multicolumn{1}{p{5cm}}{Q8: I would purchase this car.}  & \multirow{2.5}{*}{.90} \\
    & \multicolumn{1}{p{5cm}}{Q9: I would want to ride in this car as a passenger.}  & \\
    \hline
    
    \parbox[t]{2mm}{\multirow{1.8}{*}{\rotatebox[origin=c]{90}{\textbf{oth. exp.}}}} & \multicolumn{1}{p{5cm}}{Q10: I would want to share the road with this car if I were driving in one of the white cars.} & \multirow{1.8}{*}{--} \\
    \hline
    
    \parbox[t]{2mm}{\multirow{1.8}{*}{\rotatebox[origin=c]{90}{\textbf{oth. soc.}}}} & \multicolumn{1}{p{5cm}}{Q11: If I were a driver in one of the white cars on the road, I would judge the person riding in it} & \multirow{1.8}{*}{--} \\
    \hline
  \end{tabular}
    \par
    \bigskip
    Table 1. \textbf{User Study Questions.} We grouped Likert scale questions into six categories: safety, the passenger's social feelings, frustration, overall passenger experience, experience of another driver on the road, and the social feelings of another driver on the road.
\end{table}}

\prg{Subject Allocation}
We showed participants four scenarios in the order: stalled car, ambulance, speed, highway exit. We counterbalanced the order of the algorithms.

\prg{Hypothesis} \textit{We expect the cohesive car to receive higher overall ratings, leading a majority of people to prefer it.}

\subsection{Analysis}
\prg{Reliability}
Across the data, our scales show high enough reliability to warrant averaging ratings across the items for each scale: safety had Cronbach's $\alpha=.83$, social had Cronbach's $\alpha=.82$, frustration had Cronbach's $\alpha=.82$, and overall passenger experience had Cronbach's $\alpha=.9$.

\prg{Preferences After Seeing All Scenarios}
We performed a repeated measures ANOVA with cohesion as a factor and user id as a random effect for each of the scales. We found a significant effect for cohesion across the board. On the passenger side, cohesion improved the perception of safety ($F(1,11)=26.61$, $p<.0001$), improved the feeling of social belonging ($F(1,11)=12.09$, $p<.01$), decreased frustration ($F(1,11)=18.04$, $p<.01$), and improved the overall perception of riding experience ($F(1,11)=18.04$, $p<.01$). On the other driver side, users were more willing to share the road with the cohesive car ($F(1,11)=36.94$, $p<.0001$) and reported they would judge the passengers of the car less than that of the nominal car ($F(1,11)=8.29$, $p=.01$).

\begin{figure*}
    \includegraphics[width=\textwidth]{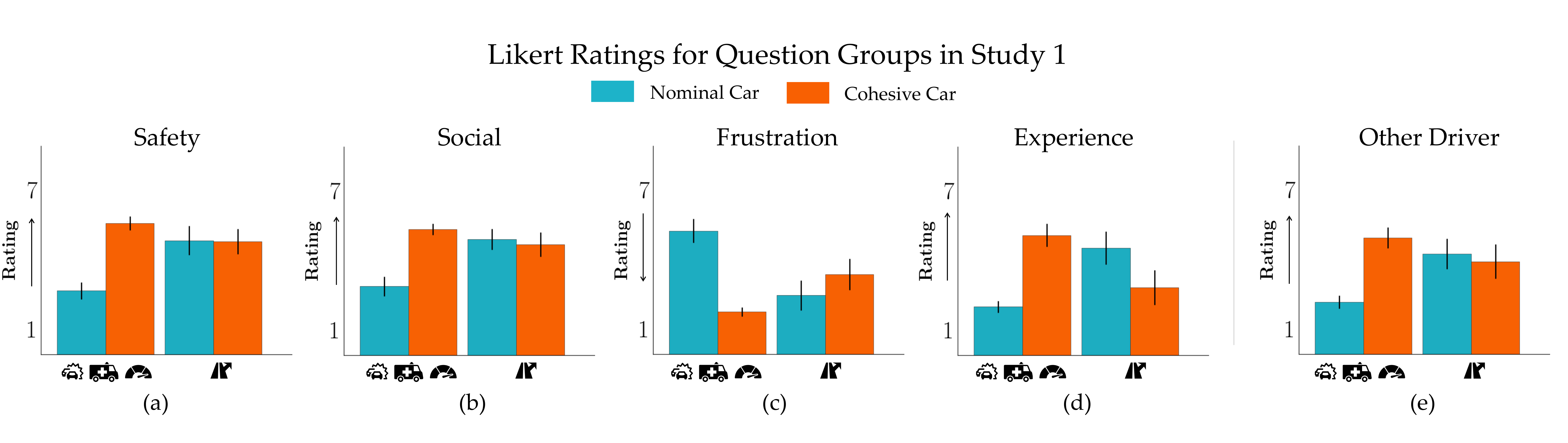}
    \caption{\textbf{Study 1 Likert Ratings.} We show the Likert ratings for the categories of safety, social feeling, frustration, overall experience, and the perspective of other drivers. We \textit{split the scenarios} into the two groups: the stalled car, ambulance and speed scenario and the exit scenario. This split corresponds to those scenarios in which the cohesive car performs well, and the exit scenario, in which it performs poorly. Higher scores are better in all cases except for frustration, where lower scores are better.}
    \label{fig:study1}
    \vspace{-0.5cm}
\end{figure*}

\prg{Preference for Individual Scenarios}
We analyzed the data for the individual scenarios by running a factorial repeated-measures ANOVA with cohesion and scenario type as factors, and user id as a random effect, for each scale. For safety, we found a significant main effect of cohesion ($F(1,77)=43.43$, $p<.0001$), but also a significant interaction effect with scenario ($F(3,77)=5.66$, $p<.01$). The post-hoc with Tukey HSD revealed that cohesion improved the perception of safety for the first three scenarios, but not for the last: this is to be expected, since in the last scenario that car mistakenly takes an exit.  All other scales yielded analogous findings, summarized in \figref{fig:study1}.

\prg{Participant Comments}
The majority of people preferred the cohesive car, although one person preferred the nominal car, describing the cohesive car as ``rigid." Many responses contained the words ``safe," ``secure" and ``predictable" when justifying a preference toward the cohesive car. One participant described the cohesive car as exhibiting ``better handling, acceleration and maneuvering capabilities ... which means [the cohesive car is]  safer to travel in."  Surprisingly, people did not dwell on the mistaken highway exit, despite our emphasis on this aspect in the prompt text. One response did describe  the cohesive car as having safety features but ``rusty," and another participant mentioned that the cohesive car may make some ``bad choices." Nevertheless, people did not invent reasons for the car exiting or place too much weight on it doing so. Crashing, on the other hand, captured the attention of participants; many mentioning that the cohesive car did not crash, whereas the nominal car did.  This led us to run a follow-up study without the the stalled car scenario, which we discuss in \sref{sec:study2}.



\section{Follow-Up: Social Cohesion for Social Settings Only}
\label{sec:study2}
In our first study, we saw that when the question is between a car that puts the passenger's life in danger versus one which accidentally takes the wrong exit, people are willing to choose the cohesive car. But what about when the benefits are much more subtle -- like keeping up with the traffic when people are going over the speed limit, or doing the right thing by stepping aside for an ambulance -- are these behaviors enough to justify taking the wrong exit? We performed a second study, in which we dropped Scenario 1 with the stalled car. 

\subsection{Experiment Design}
This study follows the format of the first one, as described in \sref{sec:study1design}.

\prg{Hypotheses}
For this second study we have two hypotheses. 

\textbf{H1}: \textit{Removing the crash scenario will lead to safety ratings which are closer between the nominal and cohesive car than in Study 1, whereas the social rating will remain similar to those in Study 1.} 

\textbf{H2}: \textit{Removing the crash scenario will cause fewer people to prefer the cohesive car.}

\subsection{Analysis}

\prg{Preferences After Seeing All Scenarios}
Our main question for this second study was what people would prefer, given that now both cars are \emph{physically} safe. We tested this via a repeated-measures ANOVA with cohesion as a factor and user id as a random effect for each of the scales. We again found significant main effect for cohesion across the board. On the passenger experience side, cohesion improved the perception of safety ($F(1,11)=8.32$, $p=.01$), improved the feeling of social belonging ($F(1,11)=12.83$, $p<.01$), decreased frustration ($F(1,11)=12.97$, $p<.01$), and improved the overall perception of riding experience ($F(1,11)=9.61$, $p=.01$). On the other driver side, users were more willing to share the road with the cohesive car ($F(1,11)=7.06$, $p=.02$) and reported they would judge the passengers of the car less ($F(1,11)=14.08$, $p=.01$). Compared to Study 1, we do see that the preferences are not as strong. In answer to ``Which car would you prefer to ride in?", 83\% of participants selected the cohesive car. In answer to ``Which car would you prefer to share the road with?", 91\% of participants selected the cohesive car.

\begin{figure}
\includegraphics[scale=0.35]{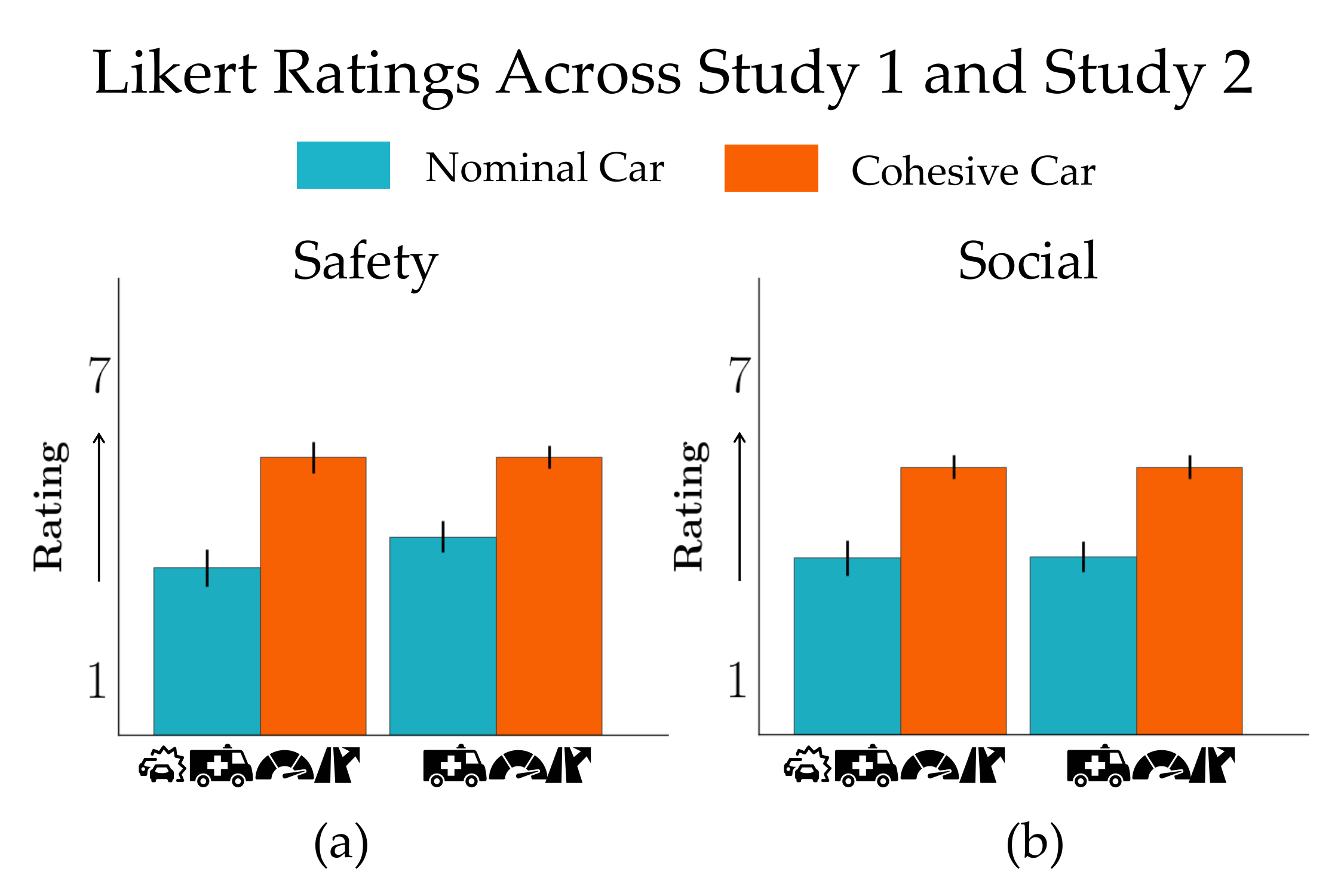}
\caption{\textbf{Likert Comparison Across Studies.} We compare the safety and social ratings across the first and second study. Notice that in the first study, the safety ratings are more disparate than in the second study, when we remove the crash scenario. The social ratings, however, remain stable between the two studies.}
\label{fig:comparison}
\vspace{-.5cm}
\end{figure}

\figref{fig:comparison} shows the key difference between the results of this study and the results of the first study. While participants still preferred the cohesive car, without the stalled car scenario the difference in safety rating for the cohesive car diminished. The difference in social rating stayed the same, which makes sense because the cohesive car retained the same social advantages, only losing its obvious safety advantage.

\prg{Preference for Individual Scenarios}
We analyzed the data for the individual scenarios by running a factorial repeated-measures ANOVA with cohesion and scenario type as factors, and user id as a random effect, for each scale. The findings were equivalent to Study 1: participants preferred the cohesive car in the first (two) scenarios, and rated it less highly in the final scenario where the cohesive car takes the unwanted exit.

\prg{Participant Comments}
This time participants used words like ``predictable" and ``considerate" to describe the cohesive car. They described the nominal car as ``oblivious" or ``unaware," and one participant stated that the car ``does not follow the social norms." Participants still mentioned safety, but when they did the distinction was less abrupt: the cohesive car only ``seems" or ``feels" safer. A few participants focused on the exit, one person explicitly calling it a ``flaw."

One participant split their preference choosing to \textit{ride in} the nominal car but \textit{share the road with} the cohesive car. Perhaps riding on a road with other drivers who follow the rules is desirable, but breaking them yourself is also attractive. Moreover, one person described the cohesive car as ``a follower," and another explicitly stated that the cohesive car ``acts more in line with unspoken norms of driving," both referenced as \textit{desirable qualities}.



\prg{Summary}
Overall, participants still chose the cohesive car (in contrast to \textbf{H2}) despite the fact that the baseline was now considered more safe (in accordance with \textbf{H1}), and despite the cohesive car erroneously exiting. These findings suggest that participants might be willing to endure the consequences of social cohesion for the benefit of riding in a car that adapts to what other drivers are doing and, in a sense, learns from their actions. 

%% file: discussion.tex
\section{Discussion}

\prg{Summary}
In this paper we presented an algorithm for an autonomous car to adapt its behavior depending on what other drivers are doing.
Our algorithm allows the robot car to learn from the behavior of other cars, and adjust its controls to match cohesive behavior.
We presented several scenarios highlighting the benefits and trade-offs associated with using our algorithm.
Our user study suggested that people are willing to accept these trade-offs in order to ride in a car which can augment its behavior when needed. 

With the same algorithm and the same parameters, we find that the car naturally adapts to what is around it.
When other cars are not cohesive, it behaves like the baseline.
But when cars are consistently doing the same thing, it switches to following that.
And, it only matches the cohesive aspects -- on a highway where everyone goes fast, with some people changing lanes and others not, it only matches the speed aspect and decides on its own what to do beyond that speed.
We also briefly looked at the sensitivity of our algorithm to noisier trajectories.
Overall, the car seamlessly integrates social cohesion into its behavior and is fairly robust. 

Of course, sometimes other cars are coincidentally cohesive and that can trick our algorithm. Despite it making mistakes like taking an exit because everyone else did, we found that people prefer it due to social benefits -- it is more ``predictable," ``considerate," and ``acts in line with unspoken norms of driving."

\prg{Limitations \& Future Work}\label{sec:limitations}
We restricted our study of this algorithm to a simple driving simulator with point-mass dynamics. We assumed perfect observations of the state of other cars and planned with a relatively short time horizon.
An implementation of our algorithm on a real car would require more sophisticated dynamics, a longer time horizon, and additional attention to safety concerns.

We picked the parameter $\beta$ to induce noticeable changes in the car's trajectory, for the purposes of analysis and our study. Future work should examine principled approaches to choosing $\beta$, the degree to which a car should adapt its behavior. Similarly, future work should perform more sensitivity analysis and examine methods to improve robustness -- the issue being that with enough features, there will always be something to follow. That might or might not be desirable.

Finally, our algorithm relies upon having a rich enough social feature representation which encodes the relevant aspects of a car's trajectory. And more fundamentally, we require the state observations of other cars on the road. Future work should further analyze the sensitivity of our algorithm to uncertainty in these features and estimates.

\prg{Conclusion} Our idea is that if everyone seems to be doing the same thing around an autonomous car, it should have an incentive to deviate from its original plan and follow along. The algorithm requires no hard-coding or explicit rules; the adaptation is autonomously learned from the trajectories of other drivers. It is perhaps surprising, and certainly elegant, that using the sample variance of straightforward features is so successful in this aim. It avoids constructing a complicated hypothesis space and is capable of handling hard-to-model aspects of the environment (for which we may not have features).  
Social cohesion is not a permanent solution, but it \emph{can help cars follow along} and better handle some aspects of the real world. 
Our work suggests that people would not only accept, but even welcome, socially-cohesive autonomous cars.




%% file: acknowledgements.tex
\section{Acknowledgements}

This work was partially supported by a Ford URP. We thank Malayandi Palan, Smitha Milli and the members of the InterACT lab for helpful discussion and feedback.